\documentclass{article}
\usepackage{ijcai24}

\usepackage{times}
\usepackage{soul}
\usepackage{url}
\usepackage[hidelinks]{hyperref}
\usepackage[utf8]{inputenc}
\usepackage[small]{caption}
\usepackage{graphicx}
\usepackage{amsmath}
\usepackage{amsthm}
\usepackage{booktabs}
\usepackage{algorithm}
\usepackage{algorithmic}
\usepackage[switch]{lineno}

\usepackage{amssymb}
\usepackage{newtxmath}
\usepackage{makecell}

\usepackage{multirow}
\usepackage[normalem]{ulem}
\useunder{\uline}{\ul}{}
\usepackage[table,xcdraw]{xcolor}
\usepackage{marvosym}
\usepackage{rotating}
\usepackage{hyperref}

\title{Sub-Adjacent Transformer: Improving Time Series Anomaly Detection \\ with Reconstruction Error from Sub-Adjacent Neighborhoods}

\author{
Wenzhen Yue$^1$
\and
Xianghua Ying$^1$\footnote{Corresponding Author}\and
Ruohao Guo$^{1}$\and
DongDong Chen$^2$\and
Ji Shi$^1$\and
Bowei Xing$^1$\and
Yuqing Zhu$^3$\And
Taiyan Chen$^1$
\affiliations
$^1$National Key Laboratory of General Artificial Intelligence, School of Intelligence Science and Technology, Peking University \\
$^2$Microsoft Cloud + AI \\ 
$^3$Tsinghua University \\
\emails
yuewenzhen@stu.pku.edu.cn,
xhying@pku.edu.cn, 	dochen@microsoft.com
}

\begin{document}

\maketitle

\begin{abstract}
    In this paper, we present the Sub-Adjacent Transformer with a novel attention mechanism for unsupervised time series anomaly detection. Unlike previous approaches that rely on all the points within some neighborhood for time point reconstruction, our method restricts the attention to regions not immediately adjacent to the target points, termed {\em sub-adjacent neighborhoods}. Our key observation is that owing to the rarity of anomalies, they typically exhibit more pronounced differences from their sub-adjacent neighborhoods than from their immediate vicinities. By focusing the attention on the sub-adjacent areas, we make the reconstruction of anomalies more challenging, thereby enhancing their detectability. Technically, our approach concentrates attention on the non-diagonal areas of the attention matrix by enlarging the corresponding elements in the training stage. To facilitate the implementation of the desired attention matrix pattern, we adopt linear attention because of its flexibility and adaptability. Moreover, a learnable mapping function is proposed to improve the performance of linear attention. Empirically, the Sub-Adjacent Transformer achieves state-of-the-art performance across six real-world anomaly detection benchmarks, covering diverse fields such as server monitoring, space exploration, and water treatment. 

\end{abstract}


\section{Introduction}

\begin{figure}[t]
   \centering
   \includegraphics[width=1\linewidth]{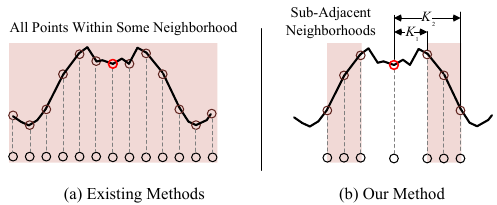}
   \caption{Illustration of our method in time domain. The point marked with a red circle, along with its neighbors, represents an anomaly on the sinusoidal signal. (a) Previous works typically utilize the attention across all the points within the window, while (b) our method encourages the use of attention of sub-adjacent neighborhoods (the highlighted area). Such an imposed constraint enlarges the reconstruction challenge for anomalies and thus improves the anomaly detection performance.}
   \label{fig1}
\end{figure}

In modern industrial systems such as data centers and smart factories, numerous sensors consistently produce significant volumes of measurements  ~\cite{zhou2019beatgan,thoc}. To effectively monitor the systems' real-time conditions and avoid potential losses, it is crucial to identify anomalies in the multivariate time series ~\cite{intro1,intro2}. This problem is known as time series anomaly detection ~\cite{acmreview:2021}. Comprehensive surveys can be found in the literature ~\cite{acmreview:2021,evaluation,ieeereview:2013}. ~\nocite{ijcai2019_ensemble}

Effective and robust time series anomaly detection remains a challenging and open problem ~\cite{intro3}. Real-world time series usually exhibit nonlinear dependencies and intricate interactions among time points. Besides, the vast scale of data makes the labeling of these anomalies time-consuming and costly in practice  ~\cite{nominality}. Therefore, time series anomaly detection is usually conducted in an unsupervised manner ~\cite{anomalytrans}, which is also the focus of this paper.

Before the era of deep learning, many classic anomaly detection methods were proposed. These include the density-estimation method proposed in ~\cite{breunig2000lof}, the clustering-based method presented in one-class SVM ~\cite{ocsvm2001}, and a series of SVDD works ~\cite{svdd2004,liu2013svdd}, along with graph-based methods ~\cite{graph2008}. Traditional methods, which use handcrafted features, struggle to accurately express the relationship among time points and often suffer from poor generalization. Deep learning-based  techniques have witnessed a diversity of methods during the recent decade. Broadly, they can be categorized into three main categories: Reconstruction-based methods reconstruct the time series and compare it with the original data ~\cite{lstm_vae_rec,rec_2021,rec_rnn_OmniAnomaly,diffusion}, whereas prediction-based methods predict future points for comparison with actual data ~\cite{gnn2021_pred,pred2,lstm-gmm-pred,nominality}. Dissimilarity-based methods emphasize the discrepancy representation between normal and abnormal points in various domains ~\cite{thoc,dissimilarity2021,anomalytrans}. It is noteworthy that these methods often combine multiple approaches to enhance performance. For example, TranAD ~\cite{tranad2022} integrates an integrated reconstruction error and discriminator loss; while ~\cite{nominality} uses prediction results as the nominal data and combines them with reconstruction results. 
~\nocite{fox1972outliers}

In this paper, we focus on the attention matrix within the Transformer module. Transformers ~\cite{transformer} have achieved great success in natural language processing ~\cite{gpt3_2020}, computer vision ~\cite{swintrans} and time series ~\cite{anomalytrans} in recent years. Previous work in \cite{anomalytrans} primarily concentrated on the dissimilarities in attention distributions. Our paper takes a different approach and introduces a simple yet effective attention learning paradigm. Our fundamental assumption is that anomalies are less related to their non-immediate neighborhoods when compared with normal points. Therefore, focusing exclusively on these non-immediate neighborhoods is likely to result in larger reconstruction errors for anomalies. Based on this assumption, our study introduces two key concepts: {\em sub-adjacent neighborhoods} and {\em sub-adjacent attention contribution}. As shown in Figure \ref{fig1}, sub-adjacent neighborhoods indicate the areas not immediately adjacent to the target point, and the sub-adjacent attention contribution is defined as the sum of particular non-diagonal elements in the corresponding column of the attention matrix. These concepts are integrated with reconstruction loss, forming the cornerstone of our anomaly detection strategy. Furthermore, we observe that the traditional {\tt Softmax} operation used in standard self-attention impedes the formation of the desired attention matrix, where the predefined non-diagonal stripes are dominant. In response, we adopt linear attention ~\cite{shen2021efficient,flattentrans} for its greater flexibility in attention matrix configurations. We also tailor the mapping function within this framework, using learnable parameters to enhance performance. The main contributions of this paper are summarized as follows.

\begin{itemize}
    \item We propose a novel attention learning regime based on the sub-adjacent neighborhoods and attention contribution. Specifically, a new attention matrix pattern is designed to enhance discrimination between anomalies and normal points. 
    \item Furthermore, we leverage the linear attention to achieve the desired attention pattern. To the best of our knowledge, this is the first introduction of linear attention with a learnable mapping function to time series anomaly detection.
    \item Extensive experiments show that the proposed Sub-Adjacent Transformer delivers state-of-the-art (SOTA) performance across six real-world benchmarks and one synthetic benchmark.      
\end{itemize}

\section{Related Works}

\paragraph{Time Series Anomaly detection.} In recent year, graph neural networks and self-attention have been explored in time series anomaly detection ~\cite{gnn2021_pred,gnn2020,Series2graph2022,anomalytrans,nominality,timesnet}. The most related work to ours is Anomaly Transformer ~\cite{anomalytrans}, which also exploits the attention information. The discrepancy between the learned Gaussian distribution and the actual distribution is used to distinguish abnormal points from normal ones. A somewhat complicated min-max training strategy is adopted to train the model. Generally, Anomaly Transformer is complex and somewhat indirect. Our proposed method is more straightforward and exhibits improved performance. ~\nocite{gnn2020,gnn2021_pred,Series2graph2022}

\paragraph{Linear Attention.} Compared with vanilla self-attention, linear attention enjoys more flexibility and lighter computation burden. Existing linear attention methods can be divided into three categories: pattern based methods, kernel based methods and mapping based methods. We mainly focus on mapping based methods in this paper. Efficient attention ~\cite{shen2021efficient} applies {\tt Softmax} function to $\mathbf{Q}$ (in a row-wise manner) and $K$ (in a column-wise manner) to ensure each row of $\mathbf{Q}\mathbf{K}^T$ sums up to 1. Hydra attention ~\cite{hydranet} studies the special case where attention heads are as many as feature dimension and $\mathbf{Q}$ and $\mathbf{K}$ are normalized so that $\left \| \mathbf{Q}_{i}  \right \|$  and $\left \| \mathbf{K}_{i}  \right \|$ share the same value. EfficientVit ~\cite{efficientvit} uses the simple ReLU function as the mapping function. FLatten Transformer ~\cite{flattentrans} proposes the power function as the mapping function to improve the focus capability. In this paper, we apply {\tt Softmax} to matrices $\mathbf{Q}$ and $\mathbf{K}$ both in a row-wise manner to approximate the focus property of the vanilla self-attention. Empirical experiments verify its performance superiority to other mapping methods in time series anomaly detection. 

\section{Methods}

\subsection{Problem Formulation}

Let $\mathbf{X} = \left \{ \mathbf{x}_1, \cdots, \mathbf{x}_T \right \}\in \mathbb{R}^{T\times D} $ denote a set of time series with $\mathbf{x}_t \in \mathbb{R}^D$, where $ T $ is the time steps and $ D $ is the number of channels. The label vector $\mathbf{y}=\left \{ y_1,\cdots, y_T\right \} $ indicates whether the corresponding time stamp is normal ($y_t=0$) or abnormal ($y_t=1$). Our task is to determine anomaly labels for all time points $ \hat{\mathbf{y} } = \left \{ \hat{y}_1, \cdots, \hat{y}_t\right \} $, where $ \hat{y}_t\in \left \{ 0,1 \right \} $, to match the ground truth $\mathbf{y}$ as much as possible. Following the practice of ~\cite{anomalytrans,evaluation,tranad2022,nominality}, we mainly focus on the F1 score with point adjustment. 

\subsection{Sub-Adjacent Neighborhoods}

Time points usually have stronger connections with their neighbors and fewer connections with distant points. This characteristic is more pronounced for anomalies ~\cite{anomalytrans}. As shown in Figure \ref{fig1}, if we rely solely on sub-adjacent neighborhoods to reconstruct time points, the reconstruction errors of anomalies will become more pronounced, thereby enhancing their distinguishability. This is the core idea of the Sub-Adjacent Transformer.

In this paper, we define the {\em sub-adjacent neighborhoods} as the region where the distance to the target point is between K1 and K2.  $K_1$ and $K_1$ are the pre-defined area bounds and satisfy $K_2\ge K_1 > 0$. The highlighted red area in Figure \ref{fig1} (b) show the sub-adjacent neighborhoods of the point marked with the red circle. 

\begin{figure}[t]
   \centering
   \includegraphics[width=0.8\linewidth]{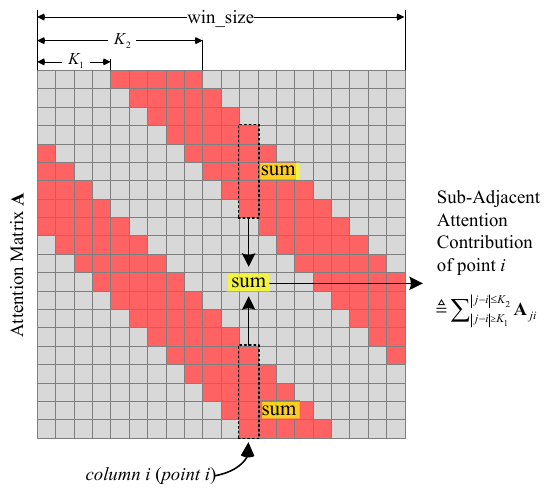}
   \caption{Illustration of attention contribution and the desired attention matrix. For clearness, only the main stripes are depicted.}
   \label{fig2}
\end{figure}

We now apply our thought to the attention matrix. The sub-adjacent neighborhoods are represented by the highlighted stripes in the attention matrix, as depicted in Figure \ref{fig2}. To focus attention on these stripes, we introduce the concept of {\em attention contribution}. This concept involves viewing the columns of the attention matrix as each point's contribution to others within the same window. Let $\mathbf{A}_{ij}$ denote the element in the $i$-th row and $j$-th column of the attention matrix $\mathbf{A}$. The value of $\mathbf{A}_{ij}$ reflects the extent of point $i$'s contribution to point $j$; the larger $\mathbf{A}_{ij}$ is, the more significant the contribution. We define the sub-adjacent attention contribution of each point as the aggregate of values within the pre-defined sub-adjacent span in the corresponding column, which is 

\begin{equation}
\begin{aligned}
    & \mathrm{SACon}\left ( \mathbf{A}  \right ) =\left [ \mathrm{SACon}\left ( \mathbf{A}_{:,i} \right )  \right ]_{i=0,\cdots,\mathrm{win\_size-1}}  \\
    & \mathrm{SACon}\left ( \mathbf{A}_{:,i} \right ) = {\textstyle \sum_{\left | j-i \right |\ge K_1 }^{\left | j-i \right |\le K_2}}\mathbf{A} _{ji}, 0\le j<win\_size
\end{aligned}
\label{SACon}
\end{equation}

\noindent where subscript $\left ( :,i \right ) $ denotes the $i$th column of the corresponding matrix, and $\mathrm{win\_size}$ is the window size. 

The sub-adjacent attention contribution plays a pivotal role in two aspects. Firstly, it steers the focus of attention towards the sub-adjacent neighborhoods by being integrated into the loss function (Eq. \ref{loss}). This is achieved by increasing $\mathrm{SACon}\left ( \mathbf{A} \right )$ for all points during the training stage. Secondly, it assists in anomaly detection. This is due to its incorporation into the anomaly score calculation, where anomalies typically show a lower attention contribution than normal points.

\begin{figure}[t]
   \centering
   \includegraphics[width=1\linewidth]{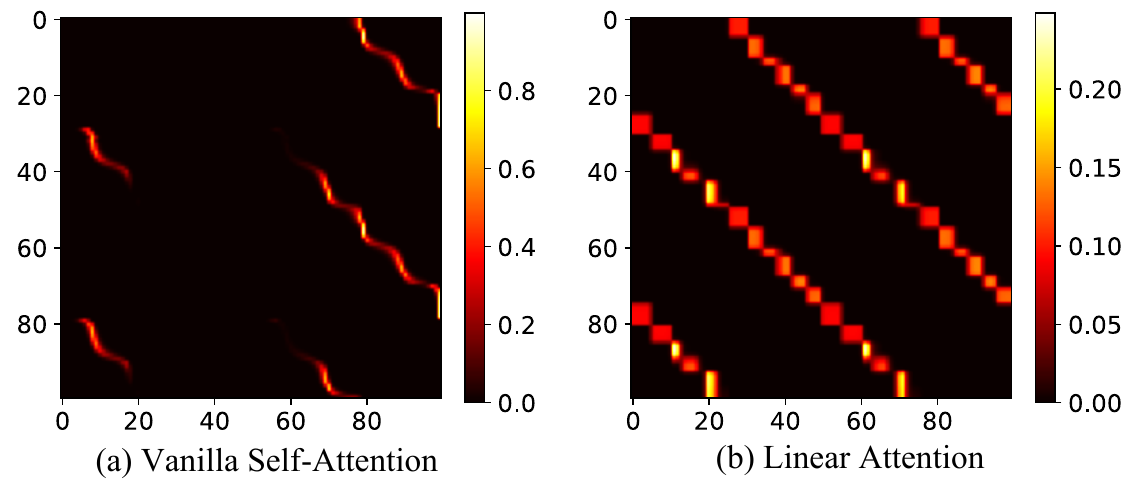}
   \caption{Attention matrices obtained using (a) vanilla self-attention and (b) the linear attention with the proposed mapping function. The SMAP dataset \protect\cite{smap} and the proposed sub-adjacent neighborhoods are used.}
   \label{fig3}
\end{figure}

Moreover, the number of the highlighted cells in Figure \ref{fig2} is lower for the marginal points (where $i<K_2$ or $i>\mathrm{ win\_size}-K_2$), leading to imbalance among points. It is mainly because the $j$ in Eq. \ref{SACon} is bounded by $0\le j<win\_size$. To break through this limitation, one plausible way is to use a circular shift function to calculate $\mathbf{A}_{ji}$ in Eq. \ref{SACon}:

\begin{equation}
    \mathbf{A} _{ji}=\left \langle \left [ \mathrm{Roll}\left ( \mathbf{Q} , i-j \right )\right ]_{i,:},  \mathbf{K} _{i,:} \right \rangle  
    \label{cyclic}
\end{equation}

\noindent where  $\mathbf{Q,K}\in \mathbb{R} ^{\mathrm{win\_size}\times {\mathrm{d_{model}}}} $  are the query and key matrix, respectively, $\left \langle \cdot,\cdot  \right \rangle $ represents the inner product of two vectors, $\mathrm{Roll}\left ( \mathbf{Q}, i-j \right ) $  cyclically shift matrix $\mathbf{Q}$ by $i-j$ along the first dimension. The $j$ values in $\mathbf{A} _{ji}$ of Eq. \ref{cyclic} can satisfy the conditions $j<0$ or $j\ge win\_size$ and ensure that the number of $j$s for each $i$ is the same. Actually, the cyclic operation in Eq. \ref{cyclic} is equivalent to setting $\mathrm{SACon}\left ( \mathbf{A}_{:,i} \right ) $ in Eq. \ref{SACon} as:

\begin{equation}
    \mathrm{SACon}\left ( \mathbf{A}_{:,i} \right ) = {\textstyle \sum_{\left | j-i \right |\ge K_1 }^{\left | j-i \right |\le K_2}}\mathbf{A} _{\left [ j \right ]  i}, \left | j \right | <win\_size
    \label{eq1_2}
\end{equation}

\noindent where $\left [ j \right ]=j+n\cdot win\_size, n\in \left \{ 0,\pm 1 \right \} $ such that $0\le \left [ j \right ]<  win\_size$. In implementation, we use Eq. \ref{eq1_2} instead of Eq. \ref{cyclic} for efficiency. As shown in Figure \ref{fig3}(b), the use of Eq. \ref{eq1_2} results in two extra side stripes in the attention matrix.

\begin{figure}[t]
   \centering
   \includegraphics[width=1\linewidth]{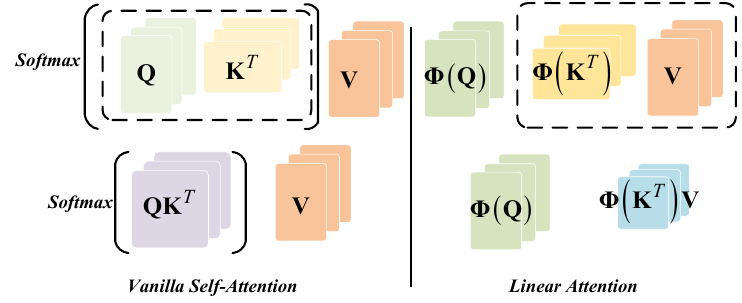}
   \caption{Illustration of vanilla self-attention and linear attention. Without the direct application of {\tt Softmax}, the attention matrix $\Phi\left (  \mathbf{Q} \right ) \Phi\left (  \mathbf{K} \right )^T$ of linear attention usually exhibits more flexibility.}
   \label{fig4}
\end{figure}

\paragraph{Linear Attention.} Vanilla self-attention employs the {\tt Softmax} function on a row-wise basis within the attention matrix, leading to competition among values in the same row, as depicted in Figure \ref{fig3}(a). In contrast, linear attention does not face such constraints. As shown in Figure \ref{fig4}, linear attention can be expressed as $\Phi\left (  \mathbf{Q} \right ) \Phi\left (  \mathbf{K} \right )^T\mathbf{V}$, where $\Phi\left (  \cdot  \right ) $ is the mapping function. As shown in Figure \ref{fig3}, linear attention demonstrates better attention matrix shaping capabilities. Quantitative results can be found in Table \ref{mappingfunc}.

The mapping function directly impacts the focusing capability and overall performance of linear attention. Herein we introduce a novel learnable mapping function:

\begin{equation}
    \Phi \left ( \cdot \right ) = \mathrm{Softmax} _{\mathrm{row}}\left (  \cdot  / \tau \right )
    \label{mymapping}
\end{equation}

\noindent where {\tt Softmax} is applied row-wise to the input matrix, and $\tau$ is a learnable parameter to adjust dynamically the {\tt Softmax} temperature. Note that in Eq. \ref{mymapping}, the {\tt Softmax} function is applied to the matrices $\mathbf{Q}$ and $\mathbf{K}$, rather than directly to the attention matrix, as is the case with vanilla self-attention. This distinction is fundamental to the increased flexibility of the attention matrix. Moreover, we set all negative values in the matrices $\mathbf{Q}$ and $\mathbf{K}$ to a large negative number, such as $-100$, to ensure that these values are close to 0 after applying the mapping function. Various mapping functions have been explored in prior research, including the power function ~\cite{flattentrans}, column-wise {\tt Softmax} ~\cite{shen2021efficient}, ReLU ~\cite{efficientvit}, and elu function ~\cite{lineartrans2020_elu}. Our empirical experiments demonstrate the effectiveness of our mapping function in time series anomaly detection, as detailed in Table \ref{mappingfunc}. 

~\nocite{diffusion2020}

\subsection{Loss Function and Anomaly Score}

\paragraph{Loss Function.} Reconstruction loss is fundamental in unsupervised time series anomaly detection. As aforementioned, we also introduce the sub-adjacent attention contribution into the loss function, which guides the model to focus on the sub-adjacent neighborhoods. By integrating these two losses, the loss function for the input series $\mathbf{X} \in \mathbb{R}^{T\times D}$ is formulated as follows:   

\begin{equation}
    \begin{aligned}
        \mathcal{L}_{\mathrm{Total}}\left ( \mathbf{X},\hat{\mathbf{X}},\mathbf{A}\right ) & = \mathcal{L}_{\mathrm{rec} } + \uplambda \cdot \mathcal{L}_{\mathrm{attn} } \\
        & = \left \|  \mathbf{X}-\hat{\mathbf{X}}\right \|_{F }^{2}-\mathrm{\uplambda}  \cdot \left \| \mathrm{SACon}\left ( \mathbf{A} \right )   \right \|_{1}   
        \label{loss}
    \end{aligned}
\end{equation}

\noindent where $\mathcal{L}_{\mathrm{rec} }$ is the reconstruction loss and $\mathcal{L}_{\mathrm{attn} }$ is the attention loss, $\uplambda >0 $ is the weight to trade off the two terms. $\left \| \cdot  \right \|  _F$ and $\left \| \cdot  \right \|  _1$ in Eq. \ref{loss} are the Frobenius and k-norm, respectively. $\hat{\mathbf{X}} \in \mathbb{R}^{T\times D}$ denotes the reconstructed $\mathbf{X}$. 

\paragraph{Anomaly Score.} To identify anomalies, we combine the sub-adjacent attention contribution with the reconstruction errors. Consistent with with the approach in ~\cite{anomalytrans}, the {\tt Softmax} function is applied to $-\mathrm{SACon}\left ( \mathbf{A} \right )$ to highlight the anomalies with less attention contribution. Subsequently, we perform an element-wise multiplication of the attention results and reconstruction errors. Finally, the anomaly score for each point can be expressed as 

\begin{equation}
    \begin{aligned}
        \mathrm{AnomalyScore} \left ( \mathbf{X}  \right ) = & \mathrm{Softmax} \left ( -\mathrm{SACon} \left ( \mathbf{A} \right )  \right ) \\ 
        & \odot \left [ \left \|\mathbf{X}_{:,i} - \hat{\mathbf{X}}_{:,i}  \right \|_{F}^{2} \right ]_{i=1,\cdots T}    
        \label{score}
    \end{aligned}
\end{equation}

\noindent where $\odot$ is the element-wise multiplication. Typically, the anomalies exhibit less attention contribution and higher  reconstruction errors, leading to larger anomaly scores. 

\paragraph{Dynamic Gaussian Scoring.} Following the practice of ~\cite{evaluation}, we fit a dynamic Gaussian distribution to anomaly scores obtained by Eq. \ref{score} and design a score based on the fitted distribution. Let $\mu _t$ and $ \sigma  _t$ denote the dynamic mean and standard variance, respectively, which can be computed in the way of sliding windows. Then the final score with dynamic Gaussian fitting can be computed via 

\begin{equation}
    \begin{aligned}
        \mathrm{DyAnoSco} _t = -\mathrm{log}  \left ( 1-\mathrm{cdf} \left ( \frac{\mathrm{AnomalyScore} _t-\mu_t}{\sigma _{t}^{2} }  \right )  \right ) 
        \label{score2}
    \end{aligned}
\end{equation}

\noindent where $\mathrm{cdf}$ is the cumulative distribution function of the standard Gaussian distribution $N\left ( 0,1 \right ) $. 

\section{Experiments}

\subsection{Datasets}

We evaluate the Sub-Adjacent Transformer on the following datasets, whose statistics are summarized in Table \ref{dataset}.

\textbf{SWaT} (Secure Water Treatment)  ~\cite{swat} is collected continuously over 11 days from 51 sensors located at a water treatment plant. \textbf{WADI} (WAter DIstribution)  ~\cite{wadi} is acquired from 123 sensors of a reduced water distribution system for 16 days.  \textbf{PSM} (Pooled Server Metrics)  ~\cite{psm} is collected from multiple servers at eBay with 26 dimensions. \textbf{MSL} (Mars Science Laboratory rover) and \textbf{SMAP} (Soil Moisture Active Passive satellite) are datasets released by NASA with 55 and 25 dimensions respectively. \textbf{SMD} (Server Machine Dataset) ~\cite{rec_rnn_OmniAnomaly} is collected from a large compute cluster, consisting of 5 weeks of data from 28 server machines with 38 sensors. \textbf{NeurIPS-TS} (NeurIPS 2021 Time Series Benchmark) is a synthetic dataset proposed by ~\cite{nipsdataset} and includes 5 kinds of anomalies that cover point- and pattern-wise behaviors: global (point), contextual (point), shapelet (pattern), seasonal (pattern), and trend (pattern). 

\begin{table}[t]
    \centering
    \begin{tabular}{llllll}
    \toprule
    Dataset & Dims & Entities & \multicolumn{1}{c}{\begin{tabular}[c]{@{}c@{}}\#Train \\ (K)\end{tabular}} & \multicolumn{1}{c}{\begin{tabular}[c]{@{}c@{}}\#Test \\ (K)\end{tabular}} & \multicolumn{1}{c}{\begin{tabular}[c]{@{}c@{}}AR \\ (\%)\end{tabular}} \\ \midrule
SWaT    & 51   & 1        & 495                                                                        & 449                                                                       & 12.14                                                                  \\
WADI    & 123  & 1        & 1209                                                                       & 172                                                                       & 5.71                                                                   \\
PSM     & 25   & 1        & 132                                                                        & 87                                                                        & 27.76                                                                  \\
MSL     & 55   & 27       & 58                                                                         & 73                                                                        & 10.48                                                                  \\
SMAP    & 25   & 55       & 140                                                                        & 444                                                                       & 12.83                                                                  \\
SMD     & 38   & 28       & 708                                                                        & 708                                                                       & 4.16                                                                   \\
NeurIPS-TS & 1    & 1        & 20                                                                         & 20                                                                        & 22.44       \\ \bottomrule
    \end{tabular}
    \caption{Datasets used in this study. AR is short for the anomaly rate. \#Train and \#Test denote the number of the training and test time points, respectively.}
    \label{dataset}
\end{table}

\subsection{Implementation Details}

\begin{table*}[t]
\begin{center}
\renewcommand\arraystretch{1}
\begin{tabular}{c|cccccc|cccccc}
\toprule
Category                                                       & \multicolumn{6}{c|}{Single-Entity}                                                                                                      & \multicolumn{6}{c}{Multi-Entity$\dag$}                                                                                                        \\ \midrule
Datset                                                         & \multicolumn{2}{c}{SWaT}                           & \multicolumn{2}{c}{WADI}                           & \multicolumn{2}{c|}{PSM}      & \multicolumn{2}{c}{MSL}                            & \multicolumn{2}{c}{SMAP}                           & \multicolumn{2}{c}{SMD}       \\ \midrule
Metric                                                         & AUC           & \multicolumn{1}{c|}{F1}            & AUC           & \multicolumn{1}{c|}{F1}            & AUC           & F1            & AUC           & \multicolumn{1}{c|}{F1}            & AUC           & \multicolumn{1}{c|}{F1}            & AUC           & F1            \\ \midrule 
DAGMM~\shortcite{dagmm}                                                         & 91.6          & \multicolumn{1}{c|}{85.3}          & 29.2          & \multicolumn{1}{c|}{20.9}          & 85.2          & 76.1          & 80.6          & \multicolumn{1}{c|}{70.1}          & 81.5          & \multicolumn{1}{c|}{71.2}          & 82.3          & 72.3          \\
LSTM-VAE~\shortcite{lstm_vae_rec}                                                     & 88.3          & \multicolumn{1}{c|}{80.5}          & 49.8          & \multicolumn{1}{c|}{38.0}          & 88.6          & 80.9          & 91.6          & \multicolumn{1}{c|}{85.4}          & 84.8          & \multicolumn{1}{c|}{75.6}          & 88.6          & 80.8          \\
MSCRED ~\shortcite{MSCRED}                                                        & 88.5          & \multicolumn{1}{c|}{80.7}          & 49.1          & \multicolumn{1}{c|}{37.4}          & 74.3          & 62.6          & 96.6          & \multicolumn{1}{c|}{93.6}          & 92.4          & \multicolumn{1}{c|}{86.6}          & 90.8          & 84.1          \\
OmniAnomaly ~\shortcite{rec_rnn_OmniAnomaly}                                                   & 92.4          & \multicolumn{1}{c|}{86.6}          & 53.9          & \multicolumn{1}{c|}{41.7}          & 77.6          & 66.4          & 94.6          & \multicolumn{1}{c|}{90.1}          & 91.6          & \multicolumn{1}{c|}{85.4}          & 98.0          & 96.2          \\
MAD-GAN  ~\shortcite{mad-gan}                                                      & 89.0          & \multicolumn{1}{c|}{81.5}          & 67.9          & \multicolumn{1}{c|}{55.6}          & 77.1          & 65.8          & 95.5          & \multicolumn{1}{c|}{91.7}          & 92.3          & \multicolumn{1}{c|}{86.5}          & 95.4          & 91.5          \\
MTAD-GAT~\shortcite{gnn2020}                                                    & 92.0          & \multicolumn{1}{c|}{86.0}          & 72.2          & \multicolumn{1}{c|}{60.2}          & 86.6          & 78.0          & 95.0          & \multicolumn{1}{c|}{90.8}          & 94.6          & \multicolumn{1}{c|}{90.1}          & 95.0          & 90.8          \\
USAD~\shortcite{audibert2020usad}                                                           & 91.1          & \multicolumn{1}{c|}{84.6}          & 55.3          & \multicolumn{1}{c|}{43.0}          & 82.5          & 72.5          & 95.2          & \multicolumn{1}{c|}{91.1}          & 89.3          & \multicolumn{1}{c|}{81.9}          & 97.2          & 94.6          \\
THOC~\shortcite{thoc}                                                    & 93.3          & \multicolumn{1}{c|}{88.1}          & 63.1          & \multicolumn{1}{c|}{50.6}          & 94.2          & 89.5          & 96.7          & \multicolumn{1}{c|}{93.7}          & 97.5          & \multicolumn{1}{c|}{95.2}          & 66.5          & 54.1          \\
UAE~\shortcite{evaluation_ieee_2022}                                                            & 92.6          & \multicolumn{1}{c|}{86.9}          & 97.8          & \multicolumn{1}{c|}{95.7}          & 96.6          & 93.6          & 95.7          & \multicolumn{1}{c|}{92.0}          & 94.3          & \multicolumn{1}{c|}{89.6}          & {\ul 98.6}    & {\ul 97.2}    \\
GDN~\shortcite{gnn2021_pred}                                                            & 96.5          & \multicolumn{1}{c|}{93.5}          & 91.7          & \multicolumn{1}{c|}{85.5}          & 95.9          & 92.3          & 94.7          & \multicolumn{1}{c|}{90.3}          & 81.1          & \multicolumn{1}{c|}{70.8}          & 81.8          & 71.6          \\
GTA~\shortcite{gta}                                                            & 95.1          & \multicolumn{1}{c|}{91.0}          & 90.7          & \multicolumn{1}{c|}{84.0}          & 91.7          & 85.5          & 95.2          & \multicolumn{1}{c|}{91.1}          & 94.7          & \multicolumn{1}{c|}{90.4}          & 95.6          & 91.9          \\
TranAD~\shortcite{tranad2022}                                                           & 89.0          & \multicolumn{1}{c|}{81.5}          & 62.0          & \multicolumn{1}{c|}{49.5}          & 93.4          & 88.2          & 97.3          & \multicolumn{1}{c|}{94.9}          & 94.0          & \multicolumn{1}{c|}{89.2}          & 98.0          & 96.1          \\

\begin{tabular}[c]{@{}c@{}} Heuristics ~\shortcite{evaluation_ieee_2022}\end{tabular}    & {\ul 98.4}    & \multicolumn{1}{c|}{{\ul 96.9}}    & 98.2          & \multicolumn{1}{c|}{96.5}          & {\ul 99.2}    & {\ul 98.5}    & {\ul 98.2}    & \multicolumn{1}{c|}{{\ul 96.5}}    & 98.0          & \multicolumn{1}{c|}{96.1}          & 96.5          & 93.4          \\
\begin{tabular}[c]{@{}c@{}}Anomaly Transformer~\shortcite{anomalytrans}\end{tabular} & 96.9          & \multicolumn{1}{c|}{94.1}          & {\ul 98.3}    & \multicolumn{1}{c|}{{\ul 96.6}}    & 98.9          & 97.9          & 96.6          & \multicolumn{1}{c|}{93.6}          & 98.3          & \multicolumn{1}{c|}{96.7}          & 95.9          & 92.3          \\ 
NPSR~\shortcite{nominality}                                                           & 97.5          & \multicolumn{1}{c|}{95.3}          & 96.7          & \multicolumn{1}{c|}{93.8}          & 97.8          & 95.7          & 97.9          & \multicolumn{1}{c|}{96.0}          & {\ul 98.9}    & \multicolumn{1}{c|}{{\ul 97.8}}    & 91.4          & 85.0          \\
TimesNet~\shortcite{timesnet}                                                           & -          & \multicolumn{1}{c|}{92.1}          & -          & \multicolumn{1}{c|}{-}          & -          & 97.5          & -          & \multicolumn{1}{c|}{85.2}          & -    & \multicolumn{1}{c|}{71.5}    & -          & 85.8          \\
\midrule
Ours                                                           & \textbf{99.5} & \multicolumn{1}{c|}{\textbf{99.0}} & \textbf{99.7} & \multicolumn{1}{c|}{\textbf{99.3}} & \textbf{99.4} & \textbf{98.9} & \textbf{98.3} & \multicolumn{1}{c|}{\textbf{96.7}} & \textbf{99.1} & \multicolumn{1}{c|}{\textbf{98.2}} & \textbf{98.7} & \textbf{97.7} \\ \bottomrule
\end{tabular}
\caption{Quantitative results for various anomoly detection methods in the six real-world datasets. AUC means area under the ROC curve. The largest and second-largest values are highlighted with bold text and underlined text, respectively. The values in this table are as $\%$ for ease of display. \dag: For multi-entity datasets, we use a single model to train and test all entities together, posing additional challenges.}
\label{mainresults}
\end{center}
\end{table*}

\begin{table*}[t]
\centering
\begin{tabular}{@{}cccccccccc@{}}
\toprule
Methods & \begin{tabular}[c]{@{}c@{}}DAGMM\\ ~\shortcite{dagmm}\end{tabular} & \begin{tabular}[c]{@{}c@{}}LSTM-VAE\\ ~\shortcite{lstm_vae_rec}\end{tabular} & \begin{tabular}[c]{@{}c@{}}OmniAnomaly\\ ~\shortcite{rec_rnn_OmniAnomaly}\end{tabular} & \begin{tabular}[c]{@{}c@{}}THOC\\ ~\shortcite{thoc}\end{tabular} 
& \begin{tabular}[c]{@{}c@{}}Heuristics \\ ~\shortcite{evaluation_ieee_2022}\end{tabular} 
& \begin{tabular}[c]{@{}c@{}}Anomaly \\Transformer\\ ~\shortcite{anomalytrans}\end{tabular} 
& \begin{tabular}[c]{@{}c@{}}NPSR\\ ~\shortcite{nominality}\end{tabular} 
& \begin{tabular}[c]{@{}c@{}}TimesNet\\ ~\shortcite{timesnet}\end{tabular} 
& Ours          \\ \midrule
AUC     & 64.3                                              & 70.5                                                 & 72.2                                                    & 74.2                                                                                        & 85.5                                                         & {\ul 86.2 }                        & 76.7  & 79.6     & \textbf{92.4} \\
F1      & 51.8                                              & 58.4                                                 & 60.2                                                    & 62.5                                                                                   & 76.3                                                        & {\ul 77.5}                                           & 65.4   & 69.5             & \textbf{86.6} \\ \bottomrule
\end{tabular}
\caption{Anomaly detection performance in the synthetic dataset NeurIPS-TS. AUC means area under the ROC curve. The largest and second-largest values are highlighted with bold text and underlined text, respectively. The values in this table are presented in percentages.}
\label{mainresults2}
\end{table*}

Following the common practice, we adopt a non-overlapping sliding window mechanism to obtain a series of sub-series. The sliding window size is set as 100 without particular statements. $K_1$ and $K_2$ are set as 20 and 30, respectively. The choices of these parameters will be discussed later in ablation studies. The points are judged to be anomalies if their anomaly scores (Eq. \ref{score2}) are larger than a certain threshold $\delta $. In this study, following the practice of paper ~\cite{nominality,anomalytrans}, the thresholds are chosen to output the best F1 scores. The widely-used point adjustment strategy ~\cite{anomalytrans,tranad2022,evaluation_ieee_2022,timesnet} is adopted.  Note that point adjustment has its practicality: one detected anomaly point will guide system administrators to identify the entire anomalous segment. The Sub-Adjacent Transformer contains 3 layers. Specifically, we set the hidden dimension $d_{\mathrm{model}}$ as 512, and the head number as 8. The hyper-parameter $\uplambda $ as 10 in Eq. \ref{loss} to balance recognition loss and attention loss. The Adam optimizer ~\cite{adam_opt} is used with an initial learning rate of $10^{-4}$. Following ~\cite{anomalytrans}, the training is early stopped within 10 epochs with the batch size of 128. Experiments are conducted using PyTorch and one NVIDIA RTX A6000 GPU.

\subsection{Main Results}

\begin{figure*}[t]
   \centering
   \includegraphics[width=1\linewidth]{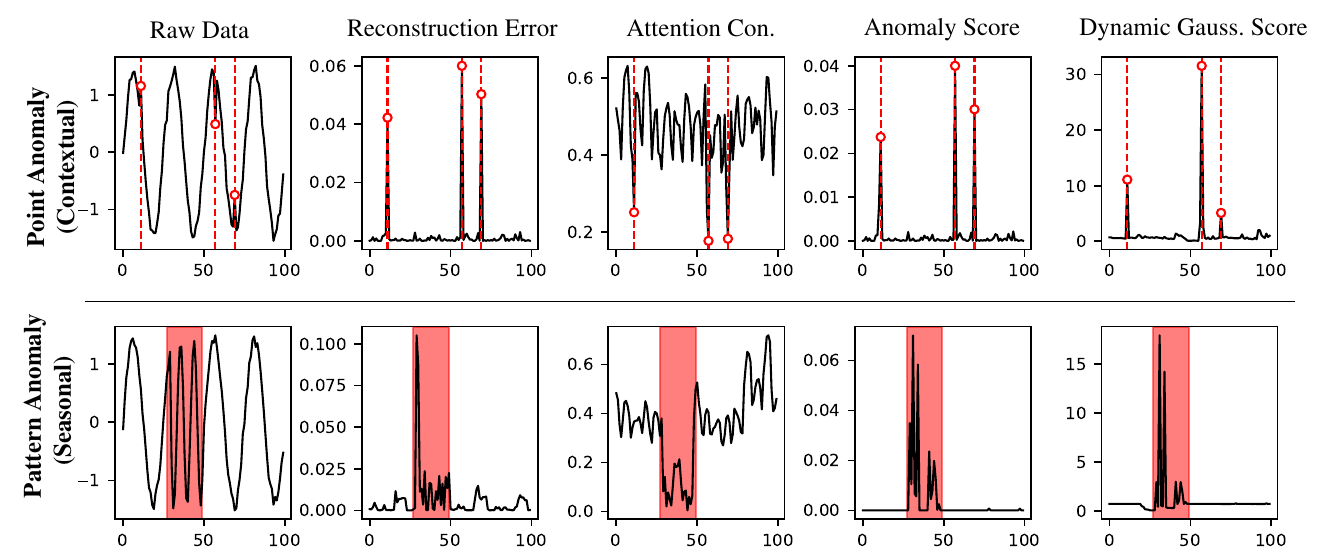}
   \caption{Visualization of detection results for different anomaly categories in NeurIPS-TS benchmark. The anomalous area are highlighted with red lines/areas. The first and second row represent point and pattern anomalies, respectively. From left to right, the columns indicate raw data, recognition error (Eq. \ref{loss}), attention contribution (Eq. \ref{SACon}), anomaly score (Eq. \ref{score}) and dynamic Gaussian score (Eq. \ref{score2}).}
   \label{fig_nips}
\end{figure*}

\begin{table}[t]
\centering
\renewcommand\arraystretch{1}
\begin{tabular}{@{}l|cccccc@{}}
\toprule
 $K_1:K_2$ & SWaT                                 & WADI                                 & PSM                                  & MSL                                  & SMAP                                 & SMD                                  \\ \midrule
 $0:0$  &   94.8        &     93.5      & 95.3                                 & 91.5                                 & 92.4                                 & 93.8                                 \\ \midrule
 $0:5$  &   96.7        &     95.9      & 97.6                                 & 93.6                                 & 93.7                                 & 94.1                                 \\ \midrule
                       $0:10$  & {\color[HTML]{FE0000} {\ul 98.5}}    & {\color[HTML]{FE0000} {\ul 98.2}}    & {\color[HTML]{FE0000} 98.7}          & 94.4                                 & 96.9                                 & 95.9                                 \\
                      $10:20$  & {\color[HTML]{FE0000} 97.6}          & {\color[HTML]{FE0000} 98.1}          & 98.1                                 & {\color[HTML]{FE0000} {\ul 96.5}}    & {\color[HTML]{FE0000} {\ul 97.8}}    & 94.4                                 \\
                      $20:30$  & {\color[HTML]{FE0000} \textbf{99.0}} & {\color[HTML]{FE0000} \textbf{99.3}} & {\color[HTML]{FE0000} \textbf{98.9}} & {\color[HTML]{FE0000} \textbf{96.7}} & {\color[HTML]{FE0000} \textbf{98.2}} & {\color[HTML]{FE0000} {\ul 97.7}}    \\
 $30:40$  & {\color[HTML]{FE0000} 97.6}          & 94.8                                 & 97.8                                 & 95.6                                 & 97.5                                 & 94.0                                 \\ \midrule
                      $10:30$  & {\color[HTML]{FE0000} 97.8}          & {\color[HTML]{FE0000} 97.8}          & {\color[HTML]{FE0000} {\ul 98.8}}    & 95.0                                 & 97.4                                 & 94.5                                 \\
 $20:40$  & {\color[HTML]{FE0000} 97.9}                                 & {\color[HTML]{FE0000} 98.0}                                 & 93.5                                 & 95.7                                 & 95.6                                 & 94.4                                 \\  \midrule
  $10:40$  & {\color[HTML]{FE0000} 98.0}          & {\color[HTML]{FE0000} 97.7}          & {\color[HTML]{FE0000} 98.7}          & 96.1                                 & 97.5                                 & {\color[HTML]{FE0000} \textbf{97.8}} \\ \bottomrule
\end{tabular}
\caption{Performance comparison with different $K_1$ and $K_2$ settings for real-world datasets. The largest value for each dataset is emphasized in bold, while the second largest value is underlined. The values are marked in red if the value is not less than SOTA.}
\label{K1andK2}
\end{table}

\begin{table*}[t]
\centering
\begin{tabular}{@{}llcccccc@{}}
\toprule
\multicolumn{2}{c}{Dataset}               & SWaT        & WADI          & PSM           & MSL           & SMAP          & SMD           \\ \midrule
\multicolumn{2}{l}{Vanilla self-attention ~\cite{transformer}}   & {\color[HTML]{FE0000} 98.6}        & {\color[HTML]{FE0000} 98.2}    & {\color[HTML]{FE0000} {\ul 98.8}}    & 94.1          & 97.4          & {\ul 96.5}    \\ \midrule
\multirow{5}{*}{\rotatebox[origin=c]{90}{Lin. Attn.}} & 
\begin{tabular}[c]{@{}l@{}}{\tt Softmax\_column} ~\cite{shen2021efficient} \end{tabular}  & {\color[HTML]{FE0000} 98.0}        & {\color[HTML]{FE0000} 97.5}          & 96.9          & {\ul 95.2}    & {\color[HTML]{FE0000} {\ul 97.8}}    & 95.8          \\ \cmidrule{2-8}
                        & \begin{tabular}[c]{@{}l@{}}Power function ~\cite{flattentrans} \end{tabular}   & {\color[HTML]{FE0000} 98.4}       & {\color[HTML]{FE0000} {\ul 98.3}}         & 98.4          & 93.7          & 97.5          & 95.9          \\ \cmidrule{2-8}
                        & \begin{tabular}[c]{@{}l@{}}ReLU ~\cite{efficientvit} \end{tabular}        & {\color[HTML]{FE0000}{\ul 98.8}}  & {\color[HTML]{FE0000}98.1}          & 97.6          & 94.3          & 97.5          & 94.8          \\ \cmidrule{2-8}
                        & \begin{tabular}[c]{@{}l@{}}ELU+1  ~\cite{lineartrans2020_elu} \end{tabular}
                                & 95.4        & 95.6          & 94.1          & 94.7          & 96.3          & 94.5          \\ \cmidrule{2-8}
                        & \begin{tabular}[c]{@{}l@{}}Ours  (Eq. \ref{mymapping})\end{tabular}            & {\color[HTML]{FE0000}\textbf{99.0}} & {\color[HTML]{FE0000}\textbf{99.3}} & {\color[HTML]{FE0000}\textbf{98.9}} & {\color[HTML]{FE0000}\textbf{96.7}} & {\color[HTML]{FE0000}\textbf{98.2}} & {\color[HTML]{FE0000}\textbf{97.7}} \\ \bottomrule
\end{tabular}
\caption{F1 values of the proposed method with vanilla self-attention and linear attention with various mapping functions. Bold and underlined text denote the largest and second-largest values per dataset. We mark the value in red if it is not less than SOTA.}
\label{mappingfunc}
\end{table*}

\begin{table*}[t]
\centering
\renewcommand\arraystretch{1.1}
\begin{tabular}{@{}lccc|llllll@{}}
\toprule
                                    & Baseline & \multicolumn{1}{l}{\begin{tabular}[c]{@{}l@{}}Linear \\ Attention\end{tabular}} & \multicolumn{1}{l|}{\begin{tabular}[c]{@{}l@{}}Dynamic \\ Scoring\end{tabular}} & SWaT & WADI & PSM  & MSL  & SMAP & SMD  \\ \midrule
\multirow{3}{*}{Ours}               &    \checkmark   &                  &                 & 98.2 & 98.0 & 98.7 & 93.9 & 97.0 & 96.1 \\
                                    &     \checkmark     &     \checkmark             &                 & $98.8_{\left ( +0.6 \right ) }$ & $99.1_{\left ( +1.1 \right ) } $ & $98.8_{\left ( +0.1 \right ) }$ & $96.4_{\left ( +2.5 \right ) }$ & $97.9_{\left ( +0.9 \right ) }$ & $97.0_{\left ( +0.9 \right ) }$ \\
                                    &      \checkmark & \checkmark               &     \checkmark       & $\mathbf{99.0_{\left ( +0.8 \right ) }} $ & $\mathbf{99.3_{\left ( +1.3 \right ) }} $ & $\mathbf{98.9_{\left ( +0.2 \right ) }} $ & $\mathbf{96.7_{\left ( +2.8 \right ) }} $ & $\mathbf{98.2_{\left ( +1.2 \right ) }} $ & $\mathbf{97.7_{\left ( +1.6 \right ) }} $ \\ \midrule
\multirow{3}{*}{\begin{tabular}[c]{@{}l@{}}Anomaly\\ Transformer\\~\shortcite{anomalytrans}\end{tabular}} &    \checkmark   &                  &                 & 94.1 & 96.6 & 97.9 & 93.6 & 96.7 & 92.3 \\
                                    &      \checkmark    &  \checkmark              &                 & $98.1_{\left ( +4.0 \right ) }$  & $97.6_{\left ( +1.0 \right ) }$  & $\mathbf{98.3_{\left ( +0.4 \right ) }} $ & $93.6_{\left ( +0.0 \right ) }$  & $97.6_{\left ( +0.9 \right ) }$  & $93.4_{\left ( +1.1 \right ) }$  \\
                                    &      \checkmark    &    \checkmark              &     \checkmark            &  $\mathbf{98.3_{\left ( +4.2 \right ) }} $ & $\mathbf{97.9_{\left ( +1.3 \right ) }} $ & $98.2_{\left ( +0.3 \right ) }$          & $\mathbf{93.7_{\left ( +0.1 \right ) }} $ & $\mathbf{97.7_{\left ( +1.0 \right ) }} $ & $\mathbf{93.6_{\left ( +1.3 \right ) }} $  \\ \bottomrule
\end{tabular}
\caption{Ablation on modules for the proposed Sub-Adjacent Transformer and Anomaly Transformer. The largest values for each dataset are highlighted with bold text. The gain with respect to baseline is given in the subscript.}
\label{module}
\end{table*}

\paragraph{Real-World Datasets.} In Table ~\ref{mainresults}, we report AUC and F1 metrics for various methods on six real-world datasets with 15 competitive baselines. The Sub-Adjacent Transformer consistently achieves state-of-the-art performance across all benchmarks, with an average improvement of 1.7 and 0.4 percentage points in F1 score over the current SOTA for single-entity and multi-entity datasets, respectively. Note that for multi-entity datasets, we combine all entities and train them together using a single model, whereas some methods, such as simple heuristics ~\cite{evaluation_ieee_2022} and NPSR ~\cite{nominality}, train each entity separately and average the results. Even though entity-to-entity variations are large (especially MSL and SMD datasets), the Sub-Adjacent Transformer still achieves better anomaly detection results using one model than the multi-model counterpart, showing the advantage of the proposed attention mechanism. We have retained all dimensions of the original datasets, even though some of them are constant and do not play a role. Additionally, we also report F1 results without using point-adjustment in the supplementary material, where the Sub-Adjacent Transformer also exhibits competitive performance.

\paragraph{Synthetic Dataset.} NeurIPS-TS is a challenging synthetic dataset. We generate the NeurIPS-TS dataset using the source codes released by ~\cite{nipsdataset}, and it includes 5 anomalous types covering both the point-wise (global and contextual) and pattern-wise (shapelet, seasonal and trend) anomalies. As shown in Table ~\ref{mainresults2}, the Sub-Adjacent Transformer achieves state-of-the-art performance with an astonishing 9.1 percentage points improvement of F1 score relative to Anomaly Transformer ~\cite{anomalytrans}, demonstrating the effectiveness of the proposed model on various anomalies. Visualization results can be found in Figure \ref{fig_nips}. We can see that anomalies do have less sub-adjacent contribution (the third column), which is consistent with our assumption. A full version of Figure \ref{fig_nips} can be found in the supplementary material. 

\subsection{Ablation Studies} 

\paragraph{Vanilla and Linear Transformer.} We compare vanilla self-attention and linear attention with different mapping functions using our method. As shown in Table \ref{mappingfunc}, the proposed {\tt Softmax}-based mapping function with learnable parameters outperforms other mapping functions and vanilla self-attention, with a maximum F1 score increase of 1.5 percentage points.

\paragraph{Choices of $K_1$ and $K_2$.} Choices of $K_1$ and $K_2$ directly affect the performance of the Sub-Adjacent Transformer. We evaluate the effect in Table \ref{K1andK2} with the window size being 100. The configuration of $K_1=20$ and $K_2=30$ typically yields the best performance (with the exception of the SMD dataset), while the performance of other settings is also satisfactory, with nearly half of the values exceeding the current SOTA. When the sub-adjacent span is getting close to the diagonal (the second and third rows in Table \ref{K1andK2}), the performance worsens. We also explore the extreme case where $K_1=K_2=0$. In that case, the performance drops significantly, highlighting the importance of sub-adjacent attention.

\begin{table}[t]
\centering
\begin{tabular}{@{}c|cccccc@{}}
\toprule
W\_S & SWaT          & WADI          & PSM           & MSL           & SMAP          & SMD           \\ \midrule
50        & 97.6          & 97.8          & 98.4          & 94.7          & 96.9          & 92.1          \\
100       & {\ul 99.0}    & \textbf{99.3} & \textbf{98.9} & \textbf{96.7} & \textbf{98.2} & \textbf{97.7} \\
150       & 98.8          & {\ul 98.8}    & {\ul 98.7}    & 94.9          & 97.3          & {\ul 96.4}    \\
200       & \textbf{99.1} & 98.7          & 98.5          & 94.9          & {\ul 97.5}    & 94.2          \\
250       & 97.2          & 98.1          & 98.3          & {\ul 95.3}    & 97.4          & 93.9          \\
300       & 98.9          & 98.2          & 98.1          & 94.3          & 96.9          & 94.3          \\ \bottomrule
\end{tabular}
\caption{F1 values for different window sizes.  The largest and second-largest values are made bold and underlined, respectively.}
\label{window_size}
\end{table}

\begin{table}[htbp]
\centering
\begin{tabular}{@{}c|cccccc@{}}
\toprule
$\uplambda$ & SWaT          & WADI          & PSM           & MSL           & SMAP          & SMD           \\ \midrule
0     & 95.8    & 94.2          & 95.6    & 92.7          & 95.3          & 94.8          \\
4     & {\ul 98.8}    & 97.8          & {\ul 98.8}    & 94.7          & 97.6          & 96.7          \\
6     & 98.6          & {\ul 98.3}    & 98.6          & {\ul 96.1}    & 97.4          & {\ul 97.3}    \\
8     & 98.3          & 97.8          & 98.7          & 95.8          & {\ul 98.0}    & 96.9          \\
10    & \textbf{99.0} & \textbf{99.3} & \textbf{98.9} & \textbf{96.7} & \textbf{98.2} & \textbf{97.7} \\
12    & 98.5          & {\ul 98.3}    & 98.6          & 95.8          & 97.6          & 97.1          \\
14    & 98.7          & 98.2          & 98.6          & 95.2          & 97.3          & 96.3          \\ \bottomrule
\end{tabular}
\caption{F1 values for different $\uplambda$ settings. Bold and underlined texts denote the largest and second largest values, respectively.}
\label{lamda_tab}
\end{table}

\paragraph{Module Ablation.} We evaluate the contribution of linear attention module and dynamic scoring module in Table \ref{module}. For the Sub-Adjacent Transformer, the baseline means the proposed sub-adjacent attention mechanism accompanied with vanilla self-attention and anomaly score (Eq. \ref{score}). Then linear attention and Gaussian dynamic scoring (Eq. \ref{score2}) are introduced in turn. As shown in Table \ref{module}, compared with the baseline, linear attention brings +1.0 improvement on average over six datasets and Gaussian dynamic scoring brings another +0.2 improvement, verifying the effectiveness of the proposed modules. \\
Furthermore, we also apply the two modules to Anomaly Transformer ~\cite{anomalytrans}. As we can see from Table \ref{module}, the proposed linear attention model provides +1.2 percentage points performance gain averagely, and Gaussian dynamic scoring provides another +0.1 gain. Moreover, as shown in Table \ref{module}, without the support of linear attention and dynamic Gaussian score, our method (baseline) still outperforms the Anomaly Transformer by an average of +1.8 percentage points, thereby validating the efficacy of our sub-adjacent attention design.

\paragraph{Choices of Window Size.} \label{winsize_para} Table \ref{window_size} gives F1 values with different window sizes for six datasets. One can see that our model exhibits good robustness to the window size. The optimal performance is typically achieved with a window size of 100, the only exception being the SWaT dataset.

\paragraph{Choices of Parameter $\uplambda$.} Table \ref{lamda_tab} compares the performance for different values of $\uplambda$. The best results are achieved when $\uplambda$ is set to 10, while the second-best results occur at different values of $\uplambda$ for different datasets. It is noteworthy that without the inclusion of attention loss ($\uplambda=0$), performance drops drastically, affirming the effectiveness of our sub-adjacent attention design.

\section{Conclusion}
In this paper, we present the Sub-Adjacent Transformer, a novel paradigm for utilizing attention in time series anomaly detection. Our method distinctively combines sub-adjacent attention contribution, linear attention and reconstruction error to effectively detect anomalies, thereby enhancing the efficacy of anomaly detection. It offers a novel perspective on the utilization of attention in this domain. Without bells and whistles, our model demonstrates superior performance across common benchmarks. We hope that the Sub-Adjacent Transformer could act as a baseline framework for the future works in time series anomaly detection. 

\section*{Acknowledgments}
This work was supported by the National Science Foundation of China (NSFC) (No. 62371009 and No. 61971008).

\bibliographystyle{named}
\bibliography{ijcai24}

\begin{thebibliography}{}

\bibitem[\protect\citeauthoryear{Abdulaal \bgroup \em et al.\egroup }{2021}]{psm}
Ahmed Abdulaal, Zhuanghua Liu, and Tomer Lancewicki.
\newblock Practical approach to asynchronous multivariate time series anomaly detection and localization.
\newblock In {\em SIGKDD}, pages 2485--2494, 2021.

\bibitem[\protect\citeauthoryear{Ahmed \bgroup \em et al.\egroup }{2017}]{wadi}
Chuadhry~Mujeeb Ahmed, Venkata~Reddy Palleti, and Aditya~P. Mathur.
\newblock Wadi: a water distribution testbed for research in the design of secure cyber physical systems.
\newblock In {\em Proceedings of the 3rd international workshop on cyber-physical systems for smart water networks}, pages 25--28, 2017.

\bibitem[\protect\citeauthoryear{Audibert \bgroup \em et al.\egroup }{2020}]{audibert2020usad}
Julien Audibert, Pietro Michiardi, Fr{\'e}d{\'e}ric Guyard, S{\'e}bastien Marti, and Maria~A. Zuluaga.
\newblock Usad: Unsupervised anomaly detection on multivariate time series.
\newblock In {\em SIGKDD}, pages 3395--3404, 2020.

\bibitem[\protect\citeauthoryear{Bl\'{a}zquez-Garc\'{\i}a \bgroup \em et al.\egroup }{2021}]{acmreview:2021}
Ane Bl\'{a}zquez-Garc\'{\i}a, Angel Conde, Usue Mori, and Jose~A Lozano.
\newblock A review on outlier/anomaly detection in time series data.
\newblock {\em ACM Computing Surveys}, 54(3):1--33, April 2021.

\bibitem[\protect\citeauthoryear{Bolya \bgroup \em et al.\egroup }{2022}]{hydranet}
Daniel Bolya, Cheng-Yang Fu, Xiaoliang Dai, Peizhao Zhang, and Judy Hoffman.
\newblock Hydra attention: Efficient attention with many heads.
\newblock In {\em ECCV}, pages 35--49. Springer, 2022.

\bibitem[\protect\citeauthoryear{Boniol and Palpanas}{2022}]{Series2graph2022}
Paul Boniol and Themis Palpanas.
\newblock Series2graph: Graph-based subsequence anomaly detection for time series.
\newblock {\em arXiv preprint arXiv:2207.12208}, 2022.

\bibitem[\protect\citeauthoryear{Breunig \bgroup \em et al.\egroup }{2000}]{breunig2000lof}
Markus~M Breunig, Hans-Peter Kriegel, Raymond~T. Ng, and J{\"o}rg Sander.
\newblock Lof: identifying density-based local outliers.
\newblock In {\em SIGMOD}, pages 93--104, 2000.

\bibitem[\protect\citeauthoryear{Brown \bgroup \em et al.\egroup }{2020}]{gpt3_2020}
Tom Brown, Benjamin Mann, Nick Ryder, Melanie Subbiah, Jared~D Kaplan, Prafulla Dhariwal, Arvind Neelakantan, Pranav Shyam, Girish Sastry, Amanda Askell, et~al.
\newblock Language models are few-shot learners.
\newblock {\em NIPS}, 33:1877--1901, 2020.

\bibitem[\protect\citeauthoryear{Cai \bgroup \em et al.\egroup }{2023}]{efficientvit}
Han Cai, Chuang Gan, and Song Han.
\newblock Efficientvit: Lightweight multi-scale attention for high-resolution dense prediction.
\newblock In {\em ICCV}, pages 17302--17313, 2023.

\bibitem[\protect\citeauthoryear{Chen \bgroup \em et al.\egroup }{2021}]{gta}
Zekai Chen, Dingshuo Chen, Xiao Zhang, Zixuan Yuan, and Xiuzhen Cheng.
\newblock Learning graph structures with transformer for multivariate time-series anomaly detection in iot.
\newblock {\em IEEE Internet of Things Journal}, 9(12):9179--9189, 2021.

\bibitem[\protect\citeauthoryear{Chen}{2018}]{intro2}
Tiankai Chen.
\newblock {\em Anomaly detection in semiconductor manufacturing through time series forecasting using neural networks}.
\newblock PhD thesis, Massachusetts Institute of Technology, 2018.

\bibitem[\protect\citeauthoryear{Cheng \bgroup \em et al.\egroup }{2008}]{graph2008}
Haibin Cheng, Pang-Ning Tan, Christopher Potter, and Steven Klooster.
\newblock A robust graph-based algorithm for detection and characterization of anomalies in noisy multivariate time series.
\newblock In {\em ICDM workshops}, pages 349--358. IEEE, 2008.

\bibitem[\protect\citeauthoryear{Deng and Hooi}{2021}]{gnn2021_pred}
Ailin Deng and Bryan Hooi.
\newblock Graph neural network-based anomaly detection in multivariate time series.
\newblock In {\em AAAI}, volume~35, pages 4027--4035, 2021.

\bibitem[\protect\citeauthoryear{Ding \bgroup \em et al.\egroup }{2019}]{lstm-gmm-pred}
Nan Ding, HaoXuan Ma, Huanbo Gao, YanHua Ma, and GuoZhen Tan.
\newblock Real-time anomaly detection based on long short-term memory and gaussian mixture model.
\newblock {\em Computers \& Electrical Engineering}, 79:106458, 2019.

\bibitem[\protect\citeauthoryear{Fox}{1972}]{fox1972outliers}
Anthony~J Fox.
\newblock Outliers in time series.
\newblock {\em Journal of the Royal Statistical Society Series B: Statistical Methodology}, 34(3):350--363, 1972.

\bibitem[\protect\citeauthoryear{Garg \bgroup \em et al.\egroup }{2022}]{evaluation_ieee_2022}
Astha Garg, Wenyu Zhang, Jules Samaran, Ramasamy Savitha, and Chuan-Sheng Foo.
\newblock An evaluation of anomaly detection and diagnosis in multivariate time series.
\newblock {\em IEEE Transactions on Neural Networks and Learning Systems}, 33(6):2508--2517, 2022.

\bibitem[\protect\citeauthoryear{Gupta \bgroup \em et al.\egroup }{2013}]{ieeereview:2013}
Manish Gupta, Jing Gao, Charu~C. Aggarwal, and Jiawei Han.
\newblock Outlier detection for temporal data: A survey.
\newblock {\em IEEE Transactions on Knowledge and data Engineering}, 26(9):2250--2267, 2013.

\bibitem[\protect\citeauthoryear{Han \bgroup \em et al.\egroup }{2023}]{flattentrans}
Dongchen Han, Xuran Pan, Yizeng Han, Shiji Song, and Gao Huang.
\newblock Flatten transformer: Vision transformer using focused linear attention.
\newblock In {\em ICCV}, pages 5961--5971, 2023.

\bibitem[\protect\citeauthoryear{Ho \bgroup \em et al.\egroup }{2020}]{diffusion2020}
Jonathan Ho, Ajay Jain, and Pieter Abbeel.
\newblock Denoising diffusion probabilistic models.
\newblock {\em NIPS}, 33:6840--6851, 2020.

\bibitem[\protect\citeauthoryear{Hundman \bgroup \em et al.\egroup }{2018a}]{pred2}
Kyle Hundman, Valentino Constantinou, Christopher Laporte, Ian Colwell, and Tom Soderstrom.
\newblock Detecting spacecraft anomalies using lstms and nonparametric dynamic thresholding.
\newblock In {\em SIGKDD}, pages 387--395, 2018.

\bibitem[\protect\citeauthoryear{Hundman \bgroup \em et al.\egroup }{2018b}]{smap}
Kyle Hundman, Valentino Constantinou, Christopher Laporte, Ian Colwell, and Tom Soderstrom.
\newblock Detecting spacecraft anomalies using lstms and nonparametric dynamic thresholding.
\newblock In {\em SIGKDD}, pages 387--395, 2018.

\bibitem[\protect\citeauthoryear{Katharopoulos \bgroup \em et al.\egroup }{2020}]{lineartrans2020_elu}
Angelos Katharopoulos, Apoorv Vyas, Nikolaos Pappas, and Fran{\c{c}}ois Fleuret.
\newblock Transformers are rnns: Fast autoregressive transformers with linear attention.
\newblock In {\em ICML}, pages 5156--5165. PMLR, 2020.

\bibitem[\protect\citeauthoryear{Kieu \bgroup \em et al.\egroup }{2019}]{ijcai2019_ensemble}
Tung Kieu, Bin Yang, Chenjuan Guo, and Christian~S. Jensen.
\newblock Outlier detection for time series with recurrent autoencoder ensembles.
\newblock In {\em IJCAI}, pages 2725--2732, 2019.

\bibitem[\protect\citeauthoryear{Kingma and Ba}{2015}]{adam_opt}
Diederik~P. Kingma and Jimmy Ba.
\newblock Adam: A method for stochastic optimization.
\newblock In {\em ICLR}, 2015.

\bibitem[\protect\citeauthoryear{Lai \bgroup \em et al.\egroup }{2021}]{nipsdataset}
Kwei-Herng Lai, Daochen Zha, Junjie Xu, Yue Zhao, Guanchu Wang, and Xia Hu.
\newblock Revisiting time series outlier detection: Definitions and benchmarks.
\newblock In {\em NeurIPS Dataset and Benchmark Track}, 2021.

\bibitem[\protect\citeauthoryear{Lai \bgroup \em et al.\egroup }{2023}]{nominality}
Chih-Yu Lai, Fan-Keng Sun, Zhengqi Gao, Jeffrey~H Lang, and Duane~S Boning.
\newblock Nominality score conditioned time series anomaly detection by point/sequential reconstruction.
\newblock {\em arXiv preprint arXiv:2310.15416}, 2023.

\bibitem[\protect\citeauthoryear{Li \bgroup \em et al.\egroup }{2019}]{mad-gan}
Dan Li, Dacheng Chen, Baihong Jin, Lei Shi, Jonathan Goh, and See-Kiong Ng.
\newblock Mad-gan: Multivariate anomaly detection for time series data with generative adversarial networks.
\newblock In {\em International conference on artificial neural networks}, pages 703--716. Springer, 2019.

\bibitem[\protect\citeauthoryear{Li \bgroup \em et al.\egroup }{2021}]{rec_2021}
Zhihan Li, Youjian Zhao, Jiaqi Han, Ya~Su, Rui Jiao, Xidao Wen, and Dan Pei.
\newblock Multivariate time series anomaly detection and interpretation using hierarchical inter-metric and temporal embedding.
\newblock In {\em SIGKDD}, pages 3220--3230, 2021.

\bibitem[\protect\citeauthoryear{Liu \bgroup \em et al.\egroup }{2013}]{liu2013svdd}
Bo~Liu, Yanshan Xiao, Longbing Cao, Zhifeng Hao, and Feiqi Deng.
\newblock Svdd-based outlier detection on uncertain data.
\newblock {\em Knowledge and information systems}, 34:597--618, 2013.

\bibitem[\protect\citeauthoryear{Liu \bgroup \em et al.\egroup }{2021}]{swintrans}
Ze~Liu, Yutong Lin, Yue Cao, Han Hu, Yixuan Wei, Zheng Zhang, Stephen Ching-Feng Lin, and Baining Guo.
\newblock Swin transformer: Hierarchical vision transformer using shifted windows.
\newblock In {\em ICCV}, pages 10012--10022, 2021.

\bibitem[\protect\citeauthoryear{Mathur and Tippenhauer}{2016}]{swat}
Aditya~P. Mathur and Nils~Ole Tippenhauer.
\newblock Swat: a water treatment testbed for research and training on ics security.
\newblock In {\em CySWATER}, pages 31--36. IEEE, 2016.

\bibitem[\protect\citeauthoryear{Park \bgroup \em et al.\egroup }{2018}]{lstm_vae_rec}
Daehyung Park, Yuuna Hoshi, and Charles~C Kemp.
\newblock A multimodal anomaly detector for robot-assisted feeding using an lstm-based variational autoencoder.
\newblock {\em IEEE Robotics and Automation Letters}, 3(3):1544--1551, 2018.

\bibitem[\protect\citeauthoryear{Pereira and Silveira}{2019}]{intro1}
João Pereira and Margarida Silveira.
\newblock Learning representations from healthcare time series data for unsupervised anomaly detection.
\newblock In {\em 2019 IEEE international conference on big data and smart computing (BigComp)}, pages 1--7. IEEE, 2019.

\bibitem[\protect\citeauthoryear{Pintilie \bgroup \em et al.\egroup }{2023}]{diffusion}
Ioana Pintilie, Andrei Manolache, and Florin Brad.
\newblock Time series anomaly detection using diffusion-based models.
\newblock {\em arXiv preprint arXiv:2311.01452}, 2023.

\bibitem[\protect\citeauthoryear{Schmidl \bgroup \em et al.\egroup }{2022}]{evaluation}
Sebastian Schmidl, Phillip Wenig, and Thorsten Papenbrock.
\newblock Anomaly detection in time series: a comprehensive evaluation.
\newblock {\em Proceedings of the VLDB Endowment}, 15(9):1779--1797, 2022.

\bibitem[\protect\citeauthoryear{Sch{\"o}lkopf \bgroup \em et al.\egroup }{2001}]{ocsvm2001}
Bernhard Sch{\"o}lkopf, John~C. Platt, J.~Shawe-Taylor, Alex Smola, and R.~C. Williamson.
\newblock Estimating the support of a high-dimensional distribution.
\newblock {\em Neural computation}, 13(7):1443--1471, 2001.

\bibitem[\protect\citeauthoryear{Shen \bgroup \em et al.\egroup }{2020}]{thoc}
Lifeng Shen, Zhuocong Li, and James~T. Kwok.
\newblock Timeseries anomaly detection using temporal hierarchical one-class network.
\newblock {\em NIPS}, 33:13016--13026, 2020.

\bibitem[\protect\citeauthoryear{Shen \bgroup \em et al.\egroup }{2021}]{shen2021efficient}
Zhuoran Shen, Mingyuan Zhang, Haiyu Zhao, Shuai Yi, , and Hongsheng Li.
\newblock Efficient attention: Attention with linear complexities.
\newblock In {\em WACV}, pages 3531--3539, 2021.

\bibitem[\protect\citeauthoryear{Su \bgroup \em et al.\egroup }{2019}]{rec_rnn_OmniAnomaly}
Ya~Su, Youjian Zhao, Chenhao Niu, Rong Liu, Wei Sun, and Dan Pei.
\newblock Robust anomaly detection for multivariate time series through stochastic recurrent neural network.
\newblock In {\em SIGKDD}, pages 2828--2837, 2019.

\bibitem[\protect\citeauthoryear{Sun \bgroup \em et al.\egroup }{2021}]{dissimilarity2021}
Fan-Keng Sun, Chris Lang, and Duane Boning.
\newblock Adjusting for autocorrelated errors in neural networks for time series.
\newblock {\em NIPS}, 34:29806--29819, 2021.

\bibitem[\protect\citeauthoryear{Tax and Duin}{2004}]{svdd2004}
David~M.J. Tax and Robert~P.W. Duin.
\newblock Support vector data description.
\newblock {\em Machine learning}, 54:45--66, 2004.

\bibitem[\protect\citeauthoryear{Tuli \bgroup \em et al.\egroup }{2022}]{tranad2022}
Shreshth Tuli, Giuliano Casale, and Nicholas~R Jennings.
\newblock Tranad: Deep transformer networks for anomaly detection in multivariate time series data.
\newblock {\em arXiv preprint arXiv:2201.07284}, 2022.

\bibitem[\protect\citeauthoryear{Vaswani \bgroup \em et al.\egroup }{2017}]{transformer}
Ashish Vaswani, Noam Shazeer, Niki Parmar, Jakob Uszkoreit, Llion Jones, Aidan~N Gomez, {\L}ukasz Kaiser, and Illia Polosukhin.
\newblock Attention is all you need.
\newblock {\em NIPS}, 30, 2017.

\bibitem[\protect\citeauthoryear{Wu \bgroup \em et al.\egroup }{2023}]{timesnet}
Haixu Wu, Tengge Hu, Yong Liu, Hang Zhou, Jianmin Wang, and Mingsheng Long.
\newblock Timesnet: Temporal 2d-variation modeling for general time series analysis.
\newblock {\em ICLR}, 2023.

\bibitem[\protect\citeauthoryear{Xu \bgroup \em et al.\egroup }{2022}]{anomalytrans}
Jiehui Xu, Haixu Wu, Jianmin Wang, and Mingsheng Long.
\newblock Anomaly transformer: Time series anomaly detection with association discrepancy.
\newblock In {\em ICLR}, 2022.

\bibitem[\protect\citeauthoryear{Zhang \bgroup \em et al.\egroup }{2019}]{MSCRED}
Chuxu Zhang, Dongjin Song, Yuncong Chen, Xinyang Feng, Cristian Lumezanu, Wei Cheng, Jingchao Ni, Bo~Zong, Haifeng Chen, and Nitesh~V Chawla.
\newblock A deep neural network for unsupervised anomaly detection and diagnosis in multivariate time series data.
\newblock In {\em AAAI}, volume~33, pages 1409--1416, 2019.

\bibitem[\protect\citeauthoryear{Zhao \bgroup \em et al.\egroup }{2020}]{gnn2020}
Hang Zhao, Yujing Wang, Juanyong Duan, Congrui Huang, Defu Cao, Yunhai Tong, Bixiong Xu, Jing Bai, Jie Tong, and Qi~Zhang.
\newblock Multivariate time-series anomaly detection via graph attention network.
\newblock In {\em ICDM}, pages 841--850. IEEE, 2020.

\bibitem[\protect\citeauthoryear{Zhao \bgroup \em et al.\egroup }{2022}]{intro3}
Yan Zhao, Liwei Deng, Xuanhao Chen, Chenjuan Guo, Bin Yang, Tung Kieu, Feiteng Huang, Torben~Bach Pedersen, Kai Zheng, and Christian~S Jensen.
\newblock A comparative study on unsupervised anomaly detection for time series: Experiments and analysis.
\newblock {\em arXiv preprint arXiv:2209.04635}, 2022.

\bibitem[\protect\citeauthoryear{Zhou \bgroup \em et al.\egroup }{2019}]{zhou2019beatgan}
Bin Zhou, Shenghua Liu, Bryan Hooi, Xueqi Cheng, and Jing Ye.
\newblock Beatgan: Anomalous rhythm detection using adversarially generated time series.
\newblock In {\em IJCAI}, volume 2019, pages 4433--4439, 2019.

\bibitem[\protect\citeauthoryear{Zong \bgroup \em et al.\egroup }{2018}]{dagmm}
Bo~Zong, Qi~Song, Martin~Renqiang Min, Wei Cheng, et~al.
\newblock Deep autoencoding gaussian mixture model for unsupervised anomaly detection.
\newblock In {\em ICLR}, 2018.

\end{thebibliography}

\appendix

\section{Overall Architecture}

\begin{figure*}[t]
   \centering
   \includegraphics[width=1\linewidth]{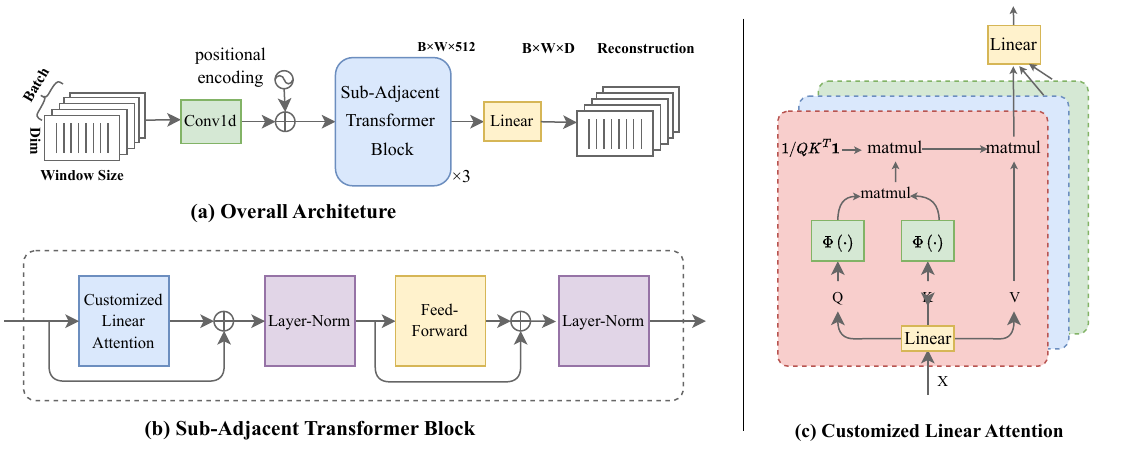}
   \caption{Network architecture for the Sub-Adjacent Transformer.}
   \label{overall_arch}
\end{figure*}

Figure \ref{overall_arch} illustrates the overall architecture of the model used in this paper. Specifically, Figure \ref{overall_arch}(c) illustrates the customized multi-head linear attention. Our main contributions lie in the sub-adjacent attention design, the introduction of linear attention with a learnable mapping function, and the corresponding loss function and anomaly score design.  

\section{Codes and Dataset Preprocessing Scripts}

It is widely recognized that the method of data preprocessing significantly influences the performance of anomaly detection. In our implementation, we use the {\tt scikit-learn} module and the {\tt StandardScaler} class to normalize the raw input data to follow a Gaussian distribution before feeding them into the network. The codes and the preprocessing script files will be released later at GitHub.

\section{F1 Without Point Adjustment}

We report the F1 scores without point adjustment in Table \ref{F1_table}. It is noticeable that our approach maintains satisfactory performance even in the absence of point adjustment. Technically, dynamic Gaussian scoring and the {\tt Softmax} operation in anomaly score are turned off, because these operations could cause the masking effect of strong values over the surrounding weak values. The reconstruction error is used as anomaly score while the loss function in the training phase is unchanged. The proposed Sub-Adjacent Transformer achieves state-of-the-art (SOTA) results in 2 out of 6 datasets and demonstrates performance closely approaching SOTA in the remaining cases, underscoring its robustness and efficacy.

\section{Full Version of Visualization}

\begin{figure*}[t]
   \centering
   \includegraphics[width=1\linewidth]{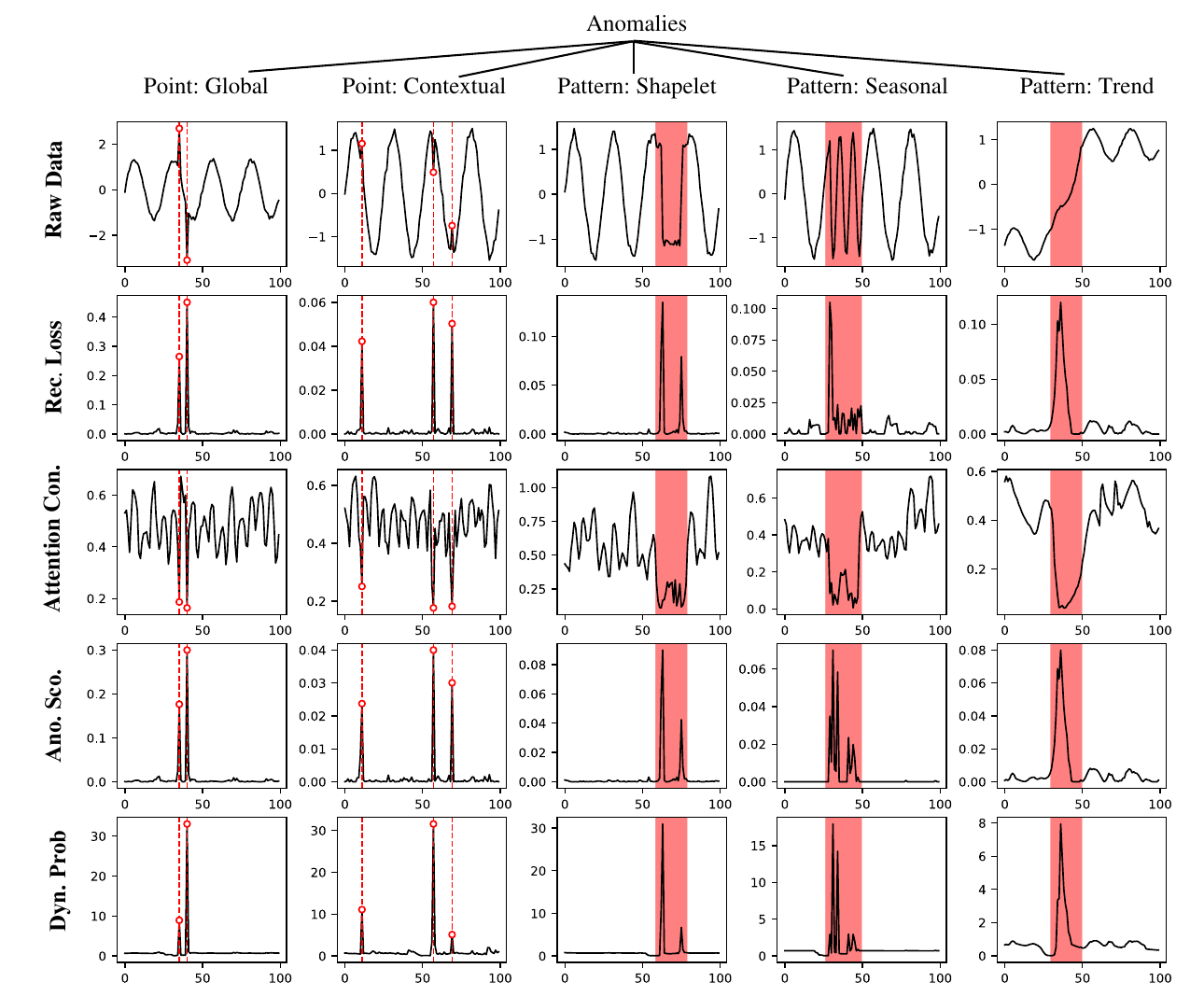}
   \caption{Visualization for different anomaly categories in the NeurIPS-TS dataset. Red lines or highlighted areas are used to mark the anomalous regions. From top to bottom, the rows are raw data, recognition loss, sub-adjacent attention contribution, anomaly score and Gaussian dynamic score, respectively. From left to right, the columns indicate global and contextual point anomalies, shapelet, seasonal and trend pattern anomalies, respectively.}
   \label{nips_visual}
\end{figure*}

NeurIPS-TS is a challenging synthetic dataset. Figure \ref{nips_visual} provides a visualization of intermediate variables that we discuss in Section 3 (Method). We can see that the Sub-adjacent Transformer can handle these 5 anomalous types well. Notably, in the third row, the attention contribution of anomalies is lower than the normal points, which is consistent with our assumption.

\section{Training Set Size}

Table \ref{trainset_ratio} reports the effects of the training set size. Training set ratio ranges from 10\% to 100\%. We can observe that the anomaly detection performance is positively correlated with the training set size and that the impact of training set size varies depending on the dataset. Particularly, for the PSM dataset, the difference between the F1 scores at 10\% and 100\% is only 1.8 percentage points. Generally speaking, our method is robust to the training set size.

\begin{table}[ht]
\centering
\renewcommand\arraystretch{1.0}
\begin{tabular}{@{}ccccccc@{}}
\toprule
T. R. & SWaT          & WADI          & PSM           & MSL           & SMAP          & SMD           \\ \midrule
10\%                                                      & 94.6          & 88.8          & 97.1          & 88.3          & 96.2          & 96.2          \\
20\%                                                      & 95.4          & 95.8          & 98.5          & 92.2          & 97.0          & 96.2          \\
40\%                                                      & 98.3          & 96.9          & 98.6          & 95.4          & 97.0          & 97.2          \\
60\%                                                      & 98.3          & 97.2          & 98.6          & 95.9          & 97.2          & 97.4          \\
80\%                                                      & {\ul 98.5}    & {\ul 98.3}    & {\ul 98.7}    & {\ul 96.2}    & {\ul 97.7}    & {\ul 97.5}    \\
100\%                                                     & \textbf{99.0} & \textbf{99.3} & \textbf{98.9} & \textbf{96.7} & \textbf{98.2} & \textbf{97.3} \\ \bottomrule
\end{tabular}
\caption{Change of F1 score with training set size. T.R. stands for the training set ratio that is actually utilized. As training data increases, the detection performance tends to increase. The effects of the training set ratio vary for different datasets due to the diverse scale of the original training sets.}
\label{trainset_ratio}
\end{table}

\section{Resource Footprint}

Table \ref{gpu_state} presents training and inference statistics for the Sub-Adjacent Transformer and its variants across six real-world datasets. The Sub-Adjacent Transformer demonstrates notable efficiency in both training and inference phases, and it maintains efficient GPU memory utilization. However, methods based on linear attention do not demonstrate significant advantages over the traditional self-attention in these respects. This phenomenon is primarily attributed to the utilization of a small window size and the implicit computation of the attention matrix. While the explicit computation of the attention matrix for small window sizes leads to enhanced efficiency, as indicated by the bracketed values in Table \ref{gpu_state}, this approach may become impractical for larger window sizes due to the resultant increase in computational complexity and memory requirements. In general, without specific statements, implicit computation is employed to align with the conventional practices of linear attention.

\begin{table*}[t]
\centering
\begin{tabular}{@{}lcccccc@{}}
\toprule
Methods             & SWaT          & WADI          & PSM           & MSL           & SMAP          & SMD           \\ \midrule
DAGMM               & 75.0          & 12.1          & 48.3          & 19.9          & 33.3          & 23.8          \\
LSTM-VAE            & 77.6          & 22.7          & 45.5          & 21.2          & 23.5          & 43.5          \\
MSCRED              & 75.7          & 4.6           & 55.6          & 25.0          & 17.0          & 38.2          \\
OmniAnomaly         & 78.2          & 22.3          & 45.2          & 20.7          & 22.7          & 47.4          \\
MAD-GAN             & 77.0          & 37.0          & 47.1          & 26.7          & 17.5          & 22.0          \\
MTAD-GAT            & 78.4          & 43.7          & 57.1          & 27.5          & 29.6          & 40.0          \\
USAD                & 79.2          & 23.3          & 47.9          & 21.1          & 22.8          & 42.6          \\
THOC                & 61.2          & 13.0          & -           & 19.0          & 24.0          & 16.8          \\
UAE                 & 45.3          & 35.4          & 42.7          & 45.1          & 39.0          & 43.5          \\
GDN                 & 81.0          & 57.0          & 55.2          & 21.7          & 25.2          & {\ul 52.9}    \\
GTA                 & 76.1          & 53.1          & 54.2          & 21.8          & 23.1          & 35.1          \\
Heuristics   & 78.9          & 35.3          & 50.9          & 23.9          & 22.9          & 49.4          \\
Anomaly Transformer & 22.0          & 10.8          & 43.4          & 19.1          & 22.7          & 8.0           \\
TranAD              & 66.9          & 41.5          & {\ul 64.9}    & 25.1          & 24.7          & 31.0          \\
NPSR                & {\ul 83.9}    & \textbf{64.2} & 64.8          & \textbf{55.1} & \textbf{51.1} & \textbf{53.5} \\ \midrule
Ours                & \textbf{84.2} & {\ul 63.5}    & \textbf{65.3} & {\ul 50.3}    & {\ul 45.2}    & 50.6          \\ \bottomrule
\end{tabular}
\caption{F1 score without point adjustment on real-world datasets. Bold text represents the best result, while underlined text represents the second best result. All channels of all the datasets are used. In our implementation, for multi-entity datasets (MSL, SMAP and SMD) we train one model for each entity and average the results. }
\label{F1_table}
\end{table*}

\begin{table*}[t]
\centering
\renewcommand\arraystretch{1.0}
\begin{tabular}{@{}llllllll@{}}
\toprule
\multirow{2}{*}{Dataset} & \multirow{2}{*}{Metric} & \multirow{2}{*}{\begin{tabular}[c]{@{}l@{}}Vanilla \\ Attention\end{tabular}} & \multicolumn{5}{c}{Linear Attention}              \\ \cmidrule(l){4-8} 
                         &                         &                                                                               & Sofmax\_col. & Pow. Func. & ReLU $\quad$  & ELU+1 $\quad$ & Ours $\quad$ \\ \midrule
\multirow{5}{*}{SWaT}    & Train. Time/Epoch (s)   & 20.6                                                                          & 34.3(30.4)   & 38.7(34.5)  & 33.6(29.3) & 34.1(30.2) & 36.7(33.1)  \\
                         & Infer. Time (s)         & 0.094                                                                         & 0.112        & 0.143       & 0.111      & 0.112      & 0.13        \\
                         & GPU Memory (G)          & 1.4                                                                           & 3.1(1.7)     & 3.6(2.2)    & 3.1(1.7)   & 3.3(1.9)   & 3.3(1.9)    \\
                         & FLOPS (M)               & 483                                                                           & 484          & 484         & 484        & 484        & 484         \\
                         & \# Params. (M)          & 4.84                                                                          & 4.84         & 4.84        & 4.84       & 4.84       & 4.84        \\ \midrule
\multirow{5}{*}{WADI}    & Train. Time/Epoch (s)   & 33.5                                                                          & 54.9(48.8)   & 61.6(55.0)  & 53.8(47.0) & 54.5(48.4) & 58.9(52.9)  \\
                         & Infer. Time (s)         & 0.047                                                                         & 0.053        & 0.065       & 0.055      & 0.054      & 0.057       \\
                         & GPU Memory (G)          & 1.5                                                                           & 3.2(1.7)     & 3.6(2.2)    & 3.2(1.7)   & 3.3(1.9)   & 3.3(1.9)    \\
                         & FLOPS (M)               & 498                                                                           & 499          & 498         & 498        & 498        & 499         \\
                         & \# Params. (M)          & 4.98                                                                          & 4.98         & 4.98        & 4.98       & 4.98       & 4.98        \\ \midrule
\multirow{5}{*}{PSM}     & Train. Time/Epoch (s)   & 55.5                                                                          & 91.4(80.8)   & 102.2(91.1) & 89.3(77.7) & 90.6(80.2) & 98.3(87.8)  \\
                         & Infer. Time (s)         & 0.018                                                                         & 0.022        & 0.026       & 0.021      & 0.021      & 0.024       \\
                         & GPU Memory (G)          & 1.45                                                                          & 3.1(1.7)     & 3.6(2.1)    & 3.2(1.7)   & 3.3(1.8)   & 3.3(1.9)    \\
                         & FLOPS (M)               & 478                                                                           & 478          & 478         & 478        & 478        & 479         \\
                         & \# Params. (M)          & 4.78                                                                          & 4.78         & 4.78        & 4.78       & 4.78       & 4.78        \\ \midrule
\multirow{5}{*}{MSL}     & Training Time (s)       & 24                                                                            & 40.2(35.8)   & 45.4(40.5)  & 39.5(34.5) & 40.1(35.6) & 42.9(38.9)  \\
                         & Infer. Time (s)         & 0.017                                                                         & 0.02         & 0.025       & 0.02       & 0.019      & 0.022       \\
                         & GPU Memory (G)          & 1.46                                                                          & 3.1(1.7)     & 3.6(2.2)    & 3.1(1.7)   & 3.3(1.8)   & 3.3(1.9)    \\
                         & FLOPS (M)               & 484                                                                           & 485          & 484         & 484        & 484        & 485         \\
                         & \# Params. (M)          & 4.84                                                                          & 4.84         & 4.84        & 4.84       & 4.84       & 4.84        \\ \midrule
\multirow{5}{*}{SMAP}    & Train. Time/Epoch (s)   & 58.1                                                                          & 96.9(85.9)   & 108.6(96.8) & 94.9(82.6) & 96.2(85.2) & 103.9(93.1) \\
                         & Infer. Time (s)         & 0.083                                                                         & 0.096        & 0.126       & 0.101      & 0.098      & 0.112       \\
                         & GPU Memory (G)          & 1.4                                                                           & 3.1(1.7)     & 3.6(2.1)    & 3.1(1.7)   & 3.2(1.8)   & 3.3(1.9)    \\
                         & FLOPS (M)               & 478                                                                           & 478          & 478         & 478        & 478        & 479         \\
                         & \# Params. (M)          & 4.78                                                                          & 4.78         & 4.78        & 4.78       & 4.78       & 4.78        \\ \midrule
\multirow{5}{*}{SMD}     & Train. Time/Epoch (s)   & 29.7                                                                          & 48.9(43.4)   & 55.1(49.1)  & 47.9(41.7) & 48.6(43.0) & 52.5(47.1)  \\
                         & Infer. Time (s)         & 0.132                                                                         & 0.157        & 0.192       & 0.158      & 0.152      & 0.184       \\
                         & GPU Memory (G)          & 1.4                                                                           & 3.1(1.7)     & 3.6(2.2)    & 3.1(1.7)   & 3.2(1.8)   & 3.3(1.9)    \\
                         & FLOPS (M)               & 481                                                                           & 481          & 481         & 481        & 481        & 482         \\
                         & \# Params. (M)          & 4.81                                                                          & 4.81         & 4.81        & 4.81       & 4.81       & 4.81        \\ \bottomrule
\end{tabular}
\caption{Training and inference statistics for our model and its variants. FLOPS and parameter numbers are obtained using the {\tt thop} module. GPU memory indicates the GPU usage during training. The values inside the brackets are obtained by explicitly calculating the attention matrix, which improves training efficiency and GPU utilization over the implicit counterpart. }
\label{gpu_state}
\end{table*}

\end{document}